\documentclass[runningheads]{llncs}

\usepackage{booktabs}
\usepackage{enumitem}
\usepackage{longtable}
\usepackage{graphicx}
\usepackage{array}
\usepackage{dcolumn}
\usepackage{amsfonts}
\usepackage{amssymb}
\usepackage{xspace}
\usepackage{xfrac}
\usepackage{array}
\usepackage{multirow}
\usepackage{xcolor}
\usepackage{footnote}
\usepackage{url}
\usepackage{makeidx}
\usepackage{epsfig}
\usepackage{amsmath}
\usepackage{multicol}
\usepackage{multirow}
\usepackage{mathtools}
\usepackage[misc]{ifsym}
\usepackage[ruled,linesnumbered]{algorithm2e}
\usepackage{tabularx}
\usepackage{array}
\usepackage{footnote}
\usepackage{subfig}
\makesavenoteenv{tabular}
\makesavenoteenv{table}
\usepackage[T1]{fontenc}
\usepackage{lmodern}
\usepackage{textcomp}

\begin{document}

\title{Forecasting with Deep Learning: Beyond Average of Average of Average Performance\thanks{This work was partially funded by projects AISym4Med (101095387) supported by Horizon Europe Cluster 1: Health, ConnectedHealth (n.º 46858), supported by Competitiveness and Internationalisation Operational Programme (POCI) and Lisbon Regional Operational Programme (LISBOA 2020), under the PORTUGAL 2020 Partnership Agreement, through the European Regional Development Fund (ERDF) and NextGenAI - Center for Responsible AI (2022-C05i0102-02), supported by IAPMEI, and also by FCT plurianual funding for 2020-2023 of LIACC (UIDB/00027/2020 UIDP/00027/2020)}}

\titlerunning{Beyond Average of Average of Average Performance}
%
\author{Vitor Cerqueira\inst{1,2} \and
Luis~Roque\inst{1,2}
\and
Carlos~Soares\inst{1,2,3}}
\authorrunning{V. Cerqueira et al.}

\institute{Faculdade de Engenharia da Universidade do Porto, Porto, Portugal \\
\email{vcerqueira@fe.up.pt}\and 
Laboratory for Artificial Intelligence and Computer Science (LIACC), Portugal \and
Fraunhofer Portugal AICOS, Portugal}

\maketitle

\begin{abstract}

Accurate evaluation of forecasting models is essential for ensuring reliable predictions.
Current practices for evaluating and comparing forecasting models focus on summarising performance into a single score, using metrics such as SMAPE.
We hypothesize that averaging performance over all samples dilutes relevant information about the relative performance of models.
Particularly, conditions in which this relative performance is different than the overall accuracy.
We address this limitation by proposing a novel framework for evaluating univariate time series forecasting models from multiple perspectives, such as one-step ahead forecasting versus multi-step ahead forecasting.
We show the advantages of this framework by comparing a state-of-the-art deep learning approach with classical forecasting techniques. While classical methods (e.g. \texttt{ARIMA}) are long-standing approaches to forecasting, deep neural networks (e.g. \texttt{NHITS}) have recently shown state-of-the-art forecasting performance in benchmark datasets. 
We conducted extensive experiments that show \texttt{NHITS} generally performs best, but its superiority varies with forecasting conditions. 
For instance, concerning the forecasting horizon, \texttt{NHITS} only outperforms classical approaches for multi-step ahead forecasting. Another relevant insight is that, when dealing with anomalies, \texttt{NHITS} is outperformed by methods such as \texttt{Theta}.
These findings highlight the importance of aspect-based model evaluation.


\end{abstract}

\section{Introduction}\label{intro}

Time series forecasting is a relevant problem in various application domains, such as finance, meteorology, or industry. The generalized interest in this task led to the development of several solutions over the past decades.
The accurate evaluation of forecasting models is essential for comparing different approaches and ensuring reliable predictions.
The typical approach for evaluating forecasts is conducted by averaging performance across all samples using metrics such as SMAPE (symmetric mean absolute percentage error) \cite{makridakis2018m4}.
As such, the estimated accuracy of a model is an average computed over multiple time steps and forecasting horizons and also across a collection of time series.

Averaging forecasting performance into a single value is convenient because it provides a simple way of quantifying the performance of models and selecting the best one among a pool of alternatives.
However, these averages dilute information that might be relevant to users. For instance, conditions in which the relative
performance of several models is different than the overall accuracy, or scenarios in which models do not behave as expected.
The real-world applicability of a model may depend on how it performs under certain conditions\footnote{Other factors may be relevant, such as computational efficiency, ease of implementation, or interpretability, but these are out of the scope of this work.} that are not captured by averaging metrics over all samples.

We address this limitation by proposing a novel framework for evaluating univariate time series forecasting models. Our approach deviates from prior works by controlling forecasting performance by various factors. We aim to uncover insights that might be obscured when error metrics are condensed into a single value. By controlling experiments across several conditions such as sampling frequency or forecasting horizon, we provide a more nuanced understanding of how different models perform under diverse scenarios. A more granular analysis enables us to pinpoint the strengths and weaknesses of different methods. This knowledge is valuable for practitioners as well as future research on forecasting methods. 


We showcase the advantages of the proposed framework by comparing a state-of-the-art deep learning approach with classical forecasting techniques.
While traditional methods such as \texttt{ARIMA} \cite{hyndman2008automatic} or exponential smoothing \cite{hyndman2008forecasting} are well-established, deep learning has recently emerged as a powerful alternative \cite{oreshkin2019n}. Several deep neural network architectures have exhibited competitive performance in benchmark competitions. These include \texttt{ES-RNN} \cite{smyl2020hybrid}, \texttt{N-BEATS} \cite{oreshkin2019n}, or \texttt{NHITS} \cite{challu2023nhits}, among others. 
The comparison of forecasting methods based on artificial neural networks with classical approaches is a topic that has been studied for a long time \cite{tang1991time,makridakis2018statistical}. 

We conduct an extensive empirical analysis comparing the deep learning approach \texttt{NHITS} \cite{challu2023nhits} with several classical forecasting methods, including \texttt{ARIMA} or \texttt{Seasonal Naive} \cite{hyndman2008automatic}. 
We select \texttt{NHITS} in particular as it has shown competitive forecasting performance with other neural networks, including \texttt{N-BEATS} \cite{oreshkin2019n}, and state-of-the-art recurrent neural networks and transformers \cite{challu2023nhits}.
We evaluate several approaches in different conditions, such as varying sampling frequency, anomalous observations, or increasing forecasting horizons. 
While \texttt{NHITS} generally performs best, its superiority varies with forecasting conditions. For instance, in terms of forecasting horizon, \texttt{NHITS} only outperforms classical approaches for multi-step ahead forecasting. When dealing with anomalies, \texttt{NHITS} is outperformed by methods such as \texttt{Theta}. 
In the interest of reproducible science, the experiments are available and fully reproducible\footnote{\url{https://github.com/vcerqueira/modelradar}}.

The rest of this paper is organized as follows. Section \ref{sec:background} provides a background to this work, including the definition of the forecasting problem and several modeling approaches used to tackle it.
In Section \ref{sec:materials}, we describe the materials and methods used in the empirical analysis carried out. The experiments and respective results are presented in Section \ref{sec:experiments} and discussed in Section \ref{sec:discussion}. We conclude the paper in Section \ref{sec:conclusions}.

\section{Background}\label{sec:background}

This section overviews several topics related to our work. We start by defining the problem and outlining a few time series models (Section \ref{sec:2.1}). 
In Section \ref{sec:2.2}, we elaborate on auto-regressive approaches focusing on how deep learning methods leverage multiple time series to build global forecasting models.
Section \ref{sec:2.3} overviews past works that compare artificial neural networks with classical approaches for univariate time series forecasting. Finally, we briefly overview evaluation practices used in forecasting problems (Section \ref{sec:2.4}).

\subsection{Time Series Forecasting}\label{sec:2.1}

A univariate time series is defined as a temporal sequence of values $Y = \{y_1, y_2, \dots,$ $y_t \}$, where $y_i \in \mathcal{Y} \subset \mathbb{R}$ is the value of $Y$ at the $i$-th timestep and $t$ is the size of $Y$. 
We address univariate time series forecasting tasks, where the goal is to predict the value of upcoming observations of the time series, $y_{t+1}, \ldots, y_{t+H}$, where $H$ denotes the forecasting horizon. 

There are several approaches to tackle this problem. One of the simplest methods is seasonal naive, which predicts the future values of a time series according to the last known observation of the same season.
\texttt{ARIMA} and exponential smoothing are two long-standing classical approaches to forecasting \cite{hyndman2018forecasting}. \texttt{ARIMA} models time series according to a linear combination of past values along with a linear combination of past errors, plus a differencing operation for integrated time series. Similarly to auto-regression, exponential smoothing models time series based on a linear combination of past observations. The simple exponential smoothing model involves a weighted average of the past values, where the weight decays exponentially as the observations are older \cite{gardner1985exponential}. 

\subsection{Forecasting with Deep Learning}\label{sec:2.2}

With machine learning, forecasting problems are framed as a supervised learning problem according to an auto-regressive type of modeling. A dataset is built using time delay embedding~\cite{bontempi2013machine}. 
Time delay embedding denotes the process of reconstructing a time series into the Euclidean space by applying sliding windows. This results in a dataset $\mathcal{D}=\{<X_{i}, y_{i}>\}^t_{i=p+1}$ where $y_i$ represents the $i$-th observation and $X_i \in \mathbb{R}^p$ is the $i$-th corresponding set of $p$ lags: $X_i = \{y_{i-1}, y_{i-2}, \dots, y_{i-p} \}$.  

Forecasting problems often involve time series databases that contain multiple univariate time series. We define a time series databases as $\mathcal{Y} = \{Y_1, Y_2, \dots, Y_n\}$, where $n$ is the number of time series in the collection. In these scenarios, forecasting approaches fall into one of two categories: local or global \cite{januschowski2020criteria}. Local methods build a model for each time series in a database. Classical forecasting techniques usually follow this approach. On the other hand, global methods train a single model using all time series in the database. 
Using several time series to train a model has been shown to lead to better forecasting performance \cite{godahewa2021ensembles}. The intuition for this effect is that the time series in a database are often related, for example, the demand time series of different related retail products. Global models can learn useful patterns in some time series that are not revealed in others, while local approaches only learn dependencies across time.

The training process of global forecasting models involves combining the data from various time series during the data preparation stage. Specifically, the training dataset $\mathcal{D}$ for a global model is composed of a concatenation of the individual datasets: $\mathcal{D} = \{\mathcal{D}_1, \dots,  \mathcal{D}_n\}$, where $\mathcal{D}_j$ is the dataset corresponding to the time series $Y_j$.
The auto-regressive formulation described above is applied to the combined dataset.

Several neural architectures have recently shown competitive forecasting performance in benchmark competitions. One of these is \texttt{NHITS} \cite{challu2023nhits}, short for Neural Hierarchical Interpolation for Time Series Forecasting. Similarly to its predecessor \texttt{N-BEATS} \cite{oreshkin2019n}, \texttt{NHITS} is based on stacks that contain blocks of multi-layer perceptrons (\texttt{MLP}) along with residual connections. The architecture behind \texttt{NHITS} also features other relevant aspects, such as multi-rate input sampling that models data with different scales or hierarchical interpolation for better long-horizon forecasting. \texttt{NHITS} has shown state-of-the-art forecasting performance relative to other deep learning approaches, including various transformers and state-of-the-art recurrent-based neural networks \cite{challu2023nhits}. Moreover, \texttt{NHITS} is significantly superior in terms of computational scalability relative to other neural-based approaches.

\subsection{Comparing Deep Learning with Classical Methods}\label{sec:2.3}

Several previous works have addressed the comparison of methods based on artificial neural networks with classical approaches for forecasting.
Hill et al. \cite{hill1996neural} pioneering work shows that \texttt{MLPs} exhibit a competitive performance with classical approaches such as \texttt{ARIMA}. Tang et al. \cite{tang1991time} also compare \texttt{MLPs} with \texttt{ARIMA}-based methods and report that \texttt{MLPs} have a competitive forecasting performance. One key finding is that the neural network performed better for long-term forecasting, while \texttt{ARIMA} was better for the short-term.

Ahmed et al. \cite{ahmed2010empirical} compare different machine learning algorithms for time series forecasting and conclude that \texttt{MLPs} and Gaussian Processes exhibit the best performance. 
In a seminal work, Makridakis et al. \cite{makridakis2018statistical} extend the study by Ahmed et al. \cite{ahmed2010empirical} by including classical approaches such as \texttt{ARIMA} or exponential smoothing. They conclude that most classical approaches, including naive, outperformed machine learning methods, including neural network algorithms. However, this study is biased towards time series dataset with a low sample size \cite{cerqueira2022case}, where neural networks become heavily over-parametrized \cite{triebe2019ar}.

The M4 forecasting competition \cite{makridakis2018m4}, which featured 100,000 from various application domains, represents an important mark for understanding the relative performance of forecasting methods. 
This competition was won by an approach called \texttt{ES-RNN} \cite{smyl2020hybrid} that combines exponential smoothing with an \texttt{LSTM} neural network trained globally.
The subsequent M5 forecasting competition \cite{makridakis2022m5} included 42,840 time series from a retail company. One of the main findings from this competition is that machine learning approaches outperformed classical methods. The winning solution was based on gradient boosting using \texttt{lightgbm} \cite{ke2017lightgbm}.

\subsection{Evaluation Metrics}\label{sec:2.4}

There are several measures to evaluate the performance of point forecasts. These fall into different categories, such as scale-dependent, scale-independent, percentage, or relative metrics. 
Hewamalage et al. \cite{hewamalage2023forecast} survey a comprehensive list of metrics and provide recommendations for which ones should be used in different scenarios. Overall, there is no consensus concerning what the best metric is. Nonetheless, for a sufficiently large sample size, most metrics agree on what the best forecasting model is \cite{koutsandreas2022selection,cerqueira2023model}.

In the benchmark M4 forecasting competition \cite{makridakis2018m4}, two metrics were used for evaluation: SMAPE and MASE (mean absolute scaled error). These are defined as follows:
\begin{equation}
    \text{SMAPE} = \frac{100\%}{n} \sum_{i=1}^{n} \frac{|\hat{y}_i - y_i|}{(|\hat{y}_i| + |y_i|)/2}
\end{equation}

\begin{equation}
    \text{MASE} = \frac{\frac{1}{n} \sum_{i=1}^{n} | y_i - \hat{y}_i |}{\frac{1}{n-m} \sum_{i=m+1}^{n} | y_i - y_{i-m} |}
\end{equation}

\noindent where $\hat{y}_i$, and $y_i$ are the forecast and actual value for the $i$-th instance, respectively, $n$ is the number of observations and $m$ is the seasonal period.
These and other metrics are usually computed across all available predictions points, which include multiple time steps, forecasting horizons, and time series.

\section{Materials and Methods}\label{sec:materials}

This section describes the materials and methods used in the experimental study. First, we present the datasets and briefly summarise their characteristics (Section \ref{sec:data}). Then, we list the forecasting methods tested in the experiments (Section \ref{sec:methods}). Then, we describe the training methodology (Section \ref{sec:evaluation}) and evaluation framework (Section \ref{sec:perfestimation}).

\subsection{Data}\label{sec:data}

We use the following benchmark datasets that were part in past forecasting competitions:
\begin{itemize}
    \item \textbf{M3} \cite{makridakis2000m3} contains a set of 3,003 time series from various application domains. The time series are split over three sampling frequencies: monthly, quarterly, and yearly;

    \item \textbf{Tourism} \cite{athanasopoulos2011tourism} contains 1,311 time series related to tourism. These also exhibit a monthly, quarterly, and yearly sampling frequency.

    \item \textbf{M4} \cite{makridakis2018m4} is a benchmark dataset with 100,000 time series from different application domains and sampling frequencies. In the interest of consistency, we use the subset of 95.000 time series that exhibit a monthly, quarterly, or yearly sampling frequency. 
\end{itemize}

Table \ref{tab:data} provides a brief summary of the datasets. Overall, the datasets contain a total of 99,140 time series with 14,898,364 observations.

\begin{table}
\caption{Summary of the datasets: number of time series, number of observations, forecast horizon, number of lags, and frequency.}
\label{tab:data}
\begin{tabular}{llr@{\hskip 0.3cm}r@{\hskip 0.3cm}r@{\hskip 0.3cm}r@{\hskip 0.3cm}r@{\hskip 0.3cm}}
\toprule
 &  & \# time series & \# observations & H & $p$ & Frequency \\
\midrule
\multirow[t]{3}{*}{M3} & Monthly & 1428 & 167562 & 18 & 23 & 12 \\
 & Quarterly & 756 & 37004 & 8 & 10 & 4 \\
 & Yearly & 645 & 18319 & 6 & 8 & 1 \\
\multirow[t]{3}{*}{M4} & Monthly & 48000 & 11246411 & 18 & 23 & 12 \\
 & Quarterly & 24000 & 2406108 & 8 & 10 & 4 \\
 & Yearly & 23000 & 858458 & 6 & 8 & 1 \\
\multirow[t]{3}{*}{Tourism} & Monthly & 366 & 109280 & 18 & 23 & 12 \\
 & Quarterly & 427 & 42544 & 8 & 10 & 4 \\
 & Yearly & 518 & 12678 & 6 & 8 & 1 \\
\midrule
Total &  & 99140 & 14898364 &  - & - & - \\
\bottomrule
\end{tabular}
\end{table}

In terms of input size\footnote{also referred to as the number of lags, or lookback window}, we follow the heuristic described by Bandara et al. \cite{bandara2020forecasting}, which leads to competitive forecasting performance \cite{leites2024lag}. 
They suggest using an input size based on the forecasting horizon and the frequency of the time series. The idea is to take the maximum value between the forecasting horizon and the frequency and then multiply the result by a factor of 1.25. We also take the ceiling to get an integer value. The resulting input size varies by sampling frequency and is reported in Table \ref{tab:data} (column $p$). We remark that this approach for selecting the input size is only adopted for deep learning. The configuration of classical approaches, such as the order of auto-regression of \texttt{ARIMA}, is selected automatically according to the process detailed in the next section.

\subsection{Methods}\label{sec:methods}

The experiments include a total of 7 forecasting approaches, 1 of which is a deep learning method. The following list describes the classical approaches:

\begin{itemize}
    \item \texttt{ARIMA}: The auto-regressive integrated moving average method that is a standard benchmark for univariate time series forecasting. The model configuration is optimized using the Akaike Information Criterion (AIC) \cite{hyndman2008automatic}; 

    \item \texttt{ETS}: The error, trend, and seasonality exponential smoothing method that is also optimized using AIC \cite{hyndman2008forecasting};

    \item \texttt{SNaive}: The seasonal naive method where forecasts are the last known observation of the same period;

    \item \texttt{RWD} (Random walk with drift) \cite{hyndman2018forecasting}: a variant of the naive method where the forecasts are adjusted according to the historical average of the time series;

    \item \texttt{SES}: The simple exponential smoothing method, with the smoothing parameter optimize by squared error minimization \cite{hyndman2008forecasting};

    \item \texttt{Theta} \cite{assimakopoulos2000theta}: The \texttt{Theta} method, with the configuration being optimized by squared error minimization.

\end{itemize}

Regarding deep learning, we include a single architecture on the experiments for conciseness. As mentioned before, we focus on \texttt{NHITS} \cite{challu2023nhits} (c.f. Section \ref{sec:2.3}), for two main reasons: i) it is significantly more computationally efficient than other architectures (50 times faster than transformers according to Challu et al. \cite{challu2023nhits}); and ii) it has shown state-of-the-art forecasting performance when compared with several other deep neural networks (e.g. \cite{challu2023nhits,cerqueira2024fly}).
We resorted to the nixtla framework\footnote{\url{https://nixtlaverse.nixtla.io/}} to implement all the above methods. Classical approaches are available on the \textit{statsforecast} Python package, while \texttt{NHITS} is implemented on \textit{neuralforecast} package.

\subsection{Training Methodology}\label{sec:evaluation}

Each classical approach follows a local methodology. On the other hand, we train \texttt{NHITS} in a global fashion according to the approach described in Section \ref{sec:2.2}. We train one \texttt{NHITS} model for each dataset listed in Table \ref{tab:data}. For instance, one model is created with all monthly time series in the M3 dataset.

We use the default configuration of \texttt{NHITS} available on \textit{neuralforecast}. Precisely, \texttt{NHITS} models are built with 3 stacks with a block of \texttt{MLPs}. Each \texttt{MLP} features 2 hidden layers, each with 512 hidden units. The activation function is set to the rectified linear unit, and \texttt{NHITS} is trained for a maximum of 1500 training steps using ADAM optimizer and a learning rate of 0.001. We use early stopping (with 50 patience steps) and model checkpointing to drive the training process.

\subsection{Evaluation Framework}\label{sec:perfestimation}

The test set is composed of the last H (one complete forecasting horizon) observations of each time series in the corresponding dataset. For example, for each monthly time series, the last 18 observations are held out for testing. 

We use SMAPE as the evaluation metric, defined in Section \ref{sec:2.4}, and apply it in three different ways:
\begin{itemize}
    \item Overall performance: The standard approach of computing forecasting performance using SMAPE on a given dataset.

    \item Expected shortfall: We use the SMAPE expected shortfall to compare different forecasting models. Expected shortfall is a financial risk measure that quantifies the expected return of a portfolio on a percentage of worst cases. We adopt this idea to our study and measure forecasting accuracy on the 5\% of time series where a given model shows the worst scores. We compute the SMAPE for each model in each time series. Then, take the average score in the 5\% of cases. This metric helps quantify and compare the models regarding their worst-case scenarios. 

    \item Win/Loss ratios: Counting how many times a model outperforms another across all time series based on SMAPE. Ratios provide a non-parametric way of comparing different models, which mitigates the effect of outliers.
\end{itemize}

These metrics are computed in different dataset conditions, specifically:
\begin{itemize}
    \item All data: Following a standard approach, we compute the metrics over all samples;

    \item Different horizons: We evaluate models in different forecasting horizons to assess if the relative performance varies across the horizon;

    \item Varying sampling frequency: We include datasets with three different sampling frequencies: monthly, quarterly, and yearly;

    \item Difficult problems: Some time series may exhibit easy-to-model patterns. In that case, an approach with a more flexible functional form, such as deep neural networks, may be unnecessary. We control for the \textit{difficulty} of a time series, which is defined in the next section.

    \item Anomalies: Finally, we analyse how models perform when forecasting anomalous observations. In this work, we consider an observation to be an anomaly if its value falls outside of the 99\% prediction interval of the \texttt{SNaive} model.
\end{itemize}

We remark that we conduct the analysis of results using all datasets jointly and not by each dataset listed in Table \ref{tab:data}.

\begin{figure}[!t]%
    \centering
    \subfloat[\centering Average SMAPE]{{\includegraphics[width=.4\textwidth]{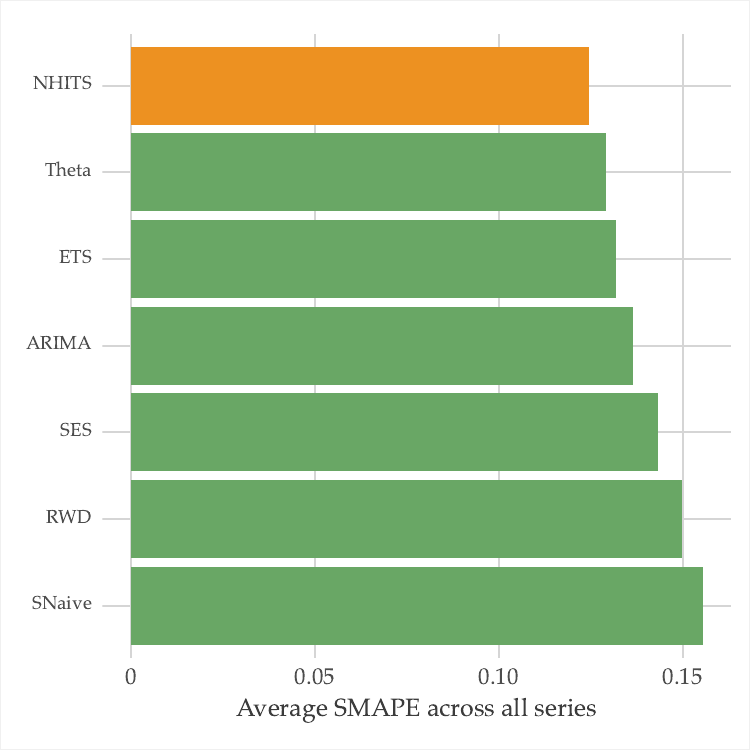} }}%
    \quad
    \subfloat[\centering SMAPE expected shortfall]{{\includegraphics[width=.4\textwidth]{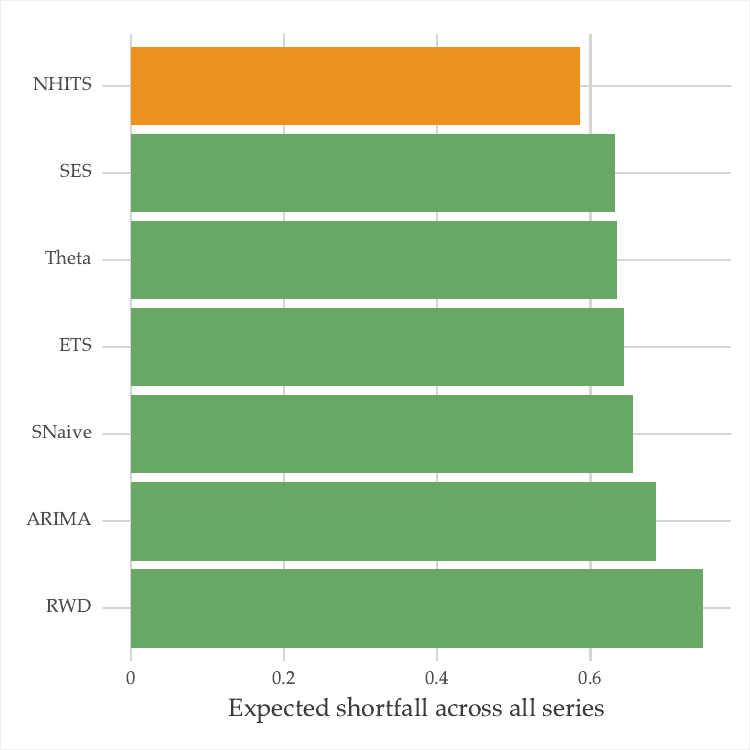}}}%
    \caption{Average SMAPE (a) and expected shortfall (b) for each model across all time series}%
    \label{fig:plot1}%
\end{figure}

\section{Experiments}\label{sec:experiments}

The evaluation framework described in the previous section is applied in a comparison of deep learning with classical forecasting techniques. In particular, the central research question posed is the following: ``How does \texttt{NHITS}, a state-of-the-art deep learning forecasting method, perform relative to classical approaches for univariate time series forecasting?" 

\begin{figure}[!ht]
    \centering
    \includegraphics[width=.9\textwidth, trim=0cm 0cm 0cm 0cm, clip=TRUE]{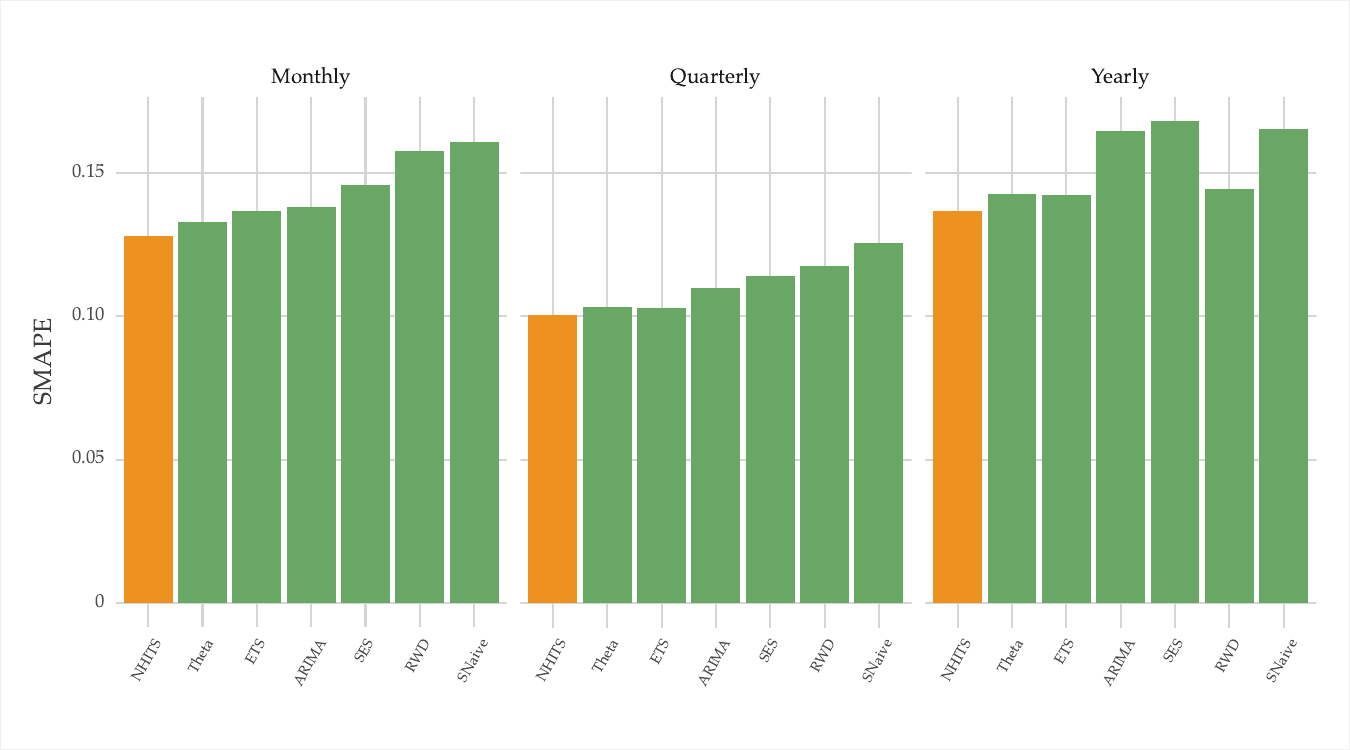}
    \caption{SMAPE scores by model and sampling frequency.}
    \label{fig:plot2}
\end{figure}

\subsection{Performance on all data}

We start by summarising forecasting performance across all time series using SMAPE. The results are shown in Figure \ref{fig:plot1}a, where \texttt{NHITS} presents the best score, outperforming all classical approaches. Among these, the Theta method exhibits the best performance. Figure \ref{fig:plot1}b shows the SMAPE expected shortfall (c.f. Section \ref{sec:perfestimation}). From a worst-case scenario perspective, \texttt{NHITS} also stands out and shows the best performance.

Then, we evaluate and compare each approach by controlling for several factors.
Figure \ref{fig:plot2} reports the SMAPE scores controlling for sampling frequency. \texttt{NHITS} shows the best performance in all cases, though the relative advantage varies in each of these. 

\begin{figure}[!t]
    \centering
    \includegraphics[width=.8\textwidth, trim=0cm 0cm 0cm 0cm, clip=TRUE]{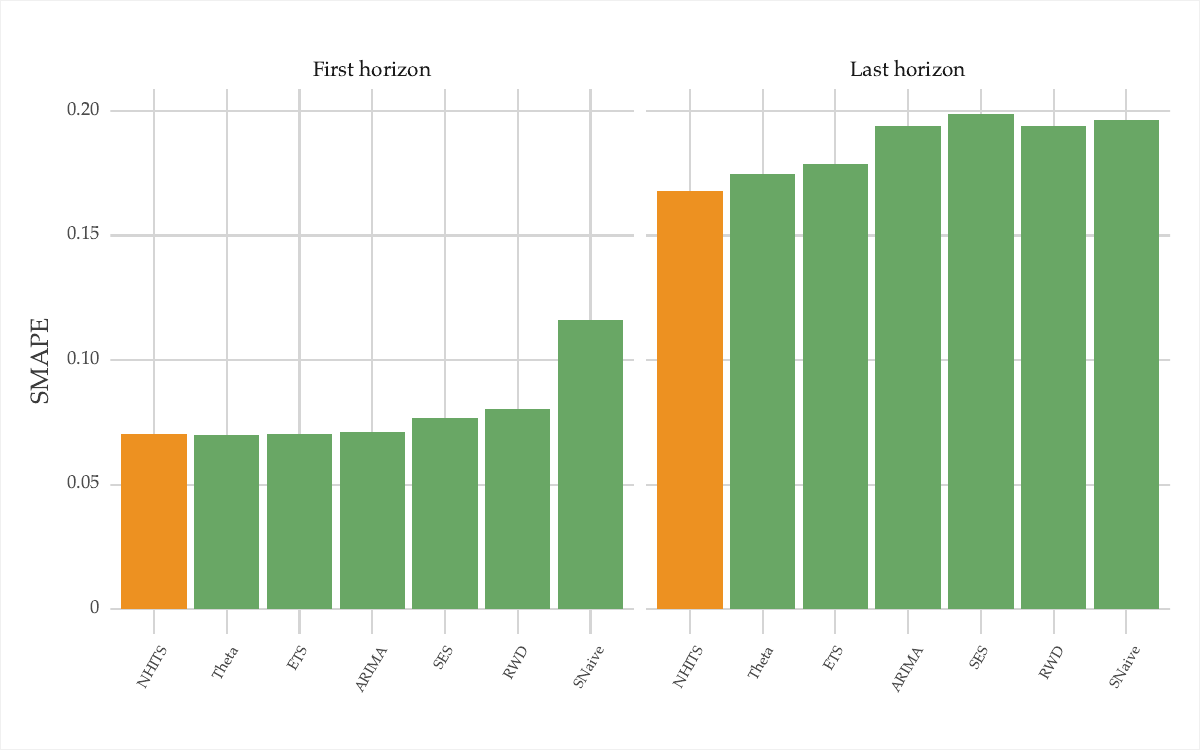}
    \caption{SMAPE scores by model and forecasting horizon.}
    \label{fig:plot3}
\end{figure}

We also controlled the experiments for forecasting horizon. We measured performance in the first and last horizon of each series, where the former equates to one-step-ahead forecasting.
The forecasting horizon varies by sampling frequency (c.f. Table \ref{tab:data}). In effect, the last horizon is different in different sampling frequencies. 
The results (Figure \ref{fig:plot3}) suggest that, for the first horizon, \texttt{NHITS} shows comparable performance with several classical approaches, such as \texttt{Theta} and \texttt{ETS}. However, in the last horizon, \texttt{NHITS} outperforms other approaches. This result is similar to the findings by Tang et al. \cite{tang1991time}, mentioned in Section \ref{sec:2.3}.

\begin{figure}%
    \centering
    \subfloat[\centering No ROPE]{{\includegraphics[width=.45\textwidth]{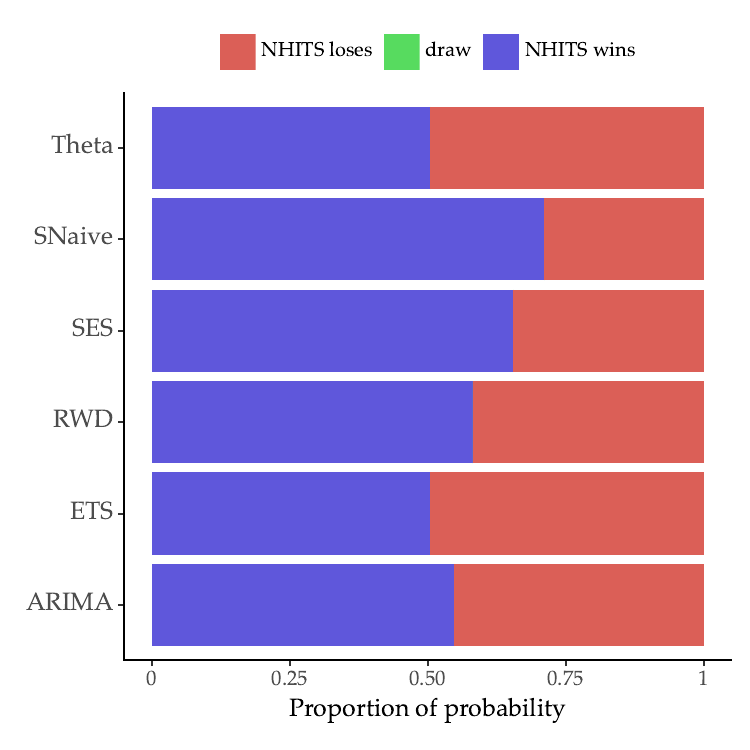} }}%
    \quad
    \subfloat[\centering ROPE=5\%]{{\includegraphics[width=.45\textwidth]{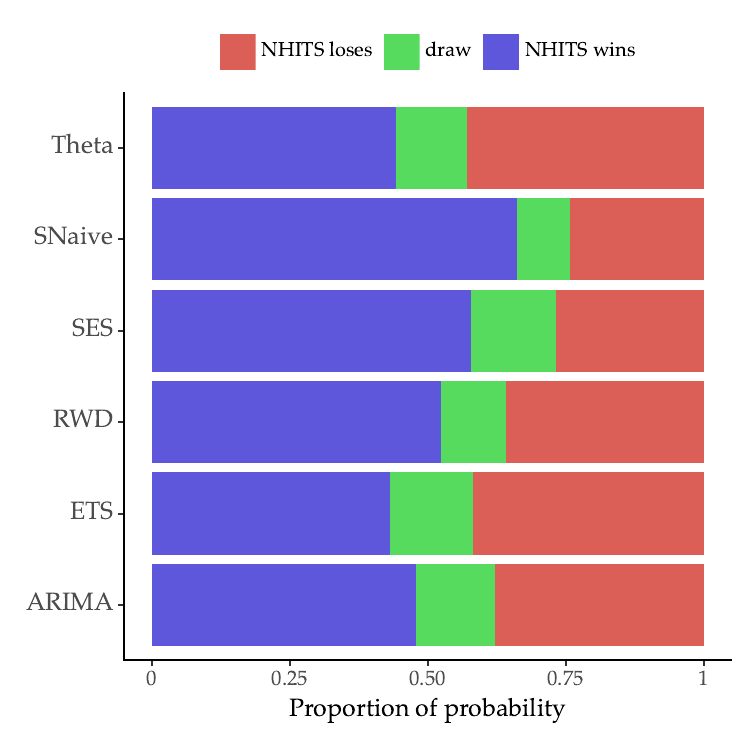} }}%
    \caption{Probability of NHITS outperforming other approaches across all time series}%
    \label{fig:plot6}%
\end{figure}

We also controlled the experiments by individual time series and computed how often \texttt{NHITS} outperformed other approaches.
Figure \ref{fig:plot6}a shows that, while \texttt{NHITS} exhibits the overall best performance, there is a reasonable chance that it is outperformed by any other method. For example, \texttt{NHITS} outperforms \texttt{Theta} in about 50\% of the 99140 time series.
We also carried out this analysis using the principles behind practical equivalence \cite{kruschke2018rejecting}. We set the region of practical equivalence (ROPE) to 5\%, so we consider two models to perform similarly if their absolute percentage difference in SMAPE is below 5\%. The results (Figure \ref{fig:plot6}b) show that \texttt{NHITS} remains competitive with all approaches in this scenario. However, there is also a reasonable chance that a given classical approach outperforms it by at least 5\%.

\subsection{Performance on difficult problems}

In the previous analysis, we considered all 99140 time series. However, some time series may exhibit patterns easily captured by a simple model. Thus, we repeat the analysis only considering difficult problems. We took a data-driven and model-based approach to define a difficult problem based on the performance of a baseline, namely \texttt{SNaive}.

\begin{figure}[ht]
    \centering
    \includegraphics[width=.9\textwidth, trim=0cm 0cm 0cm 0cm, clip=TRUE]{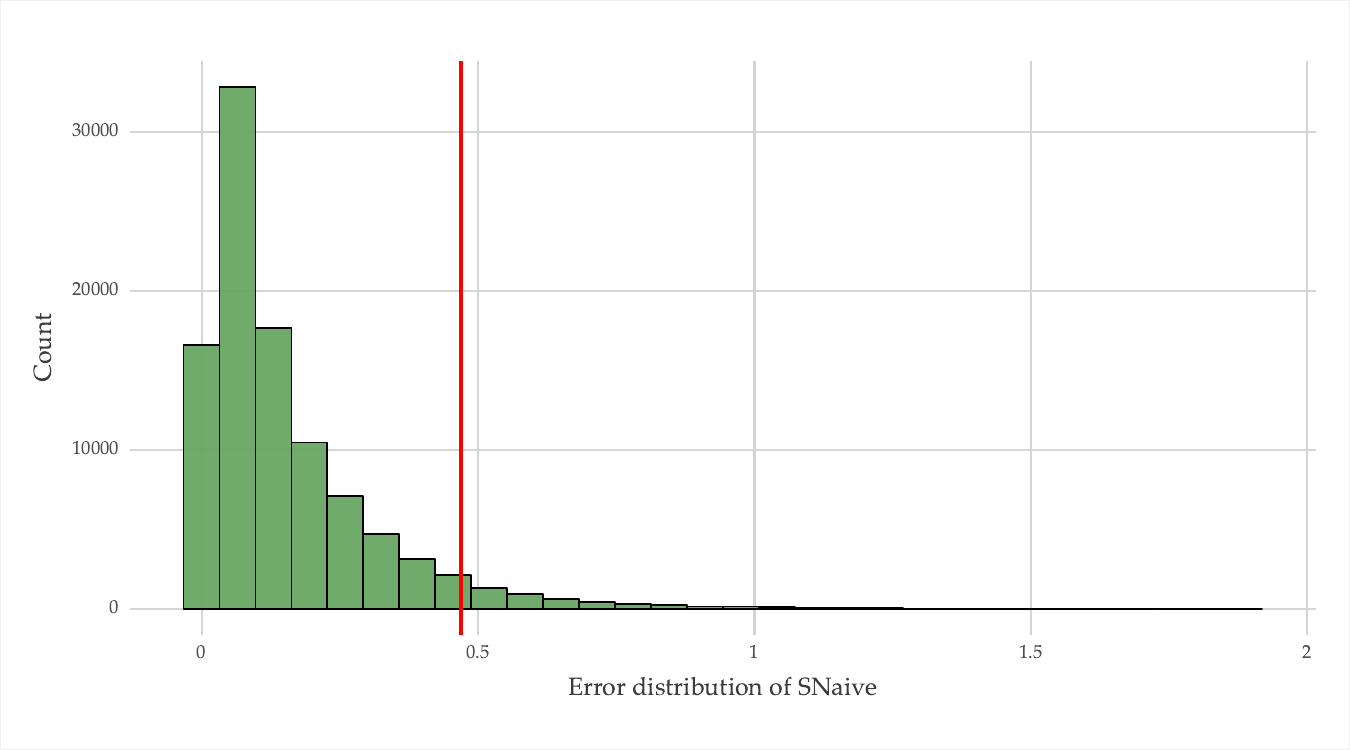}
    \caption{SMAPE distribution of SNaive across all time series. The vertical line depicts the 95\% score percentile.}
    \label{fig:plot7_0}
\end{figure}

Figure \ref{fig:plot7_0} shows the distribution of SMAPE performance by \texttt{SNaive} across all time series. The vertical line depicts the 95\% score percentile. We consider a difficult problem to be any time series corresponding to the right side of the vertical line. 

\begin{figure}%
    \centering
    \subfloat[\centering Average SMAPE]{{\includegraphics[width=.45\textwidth]{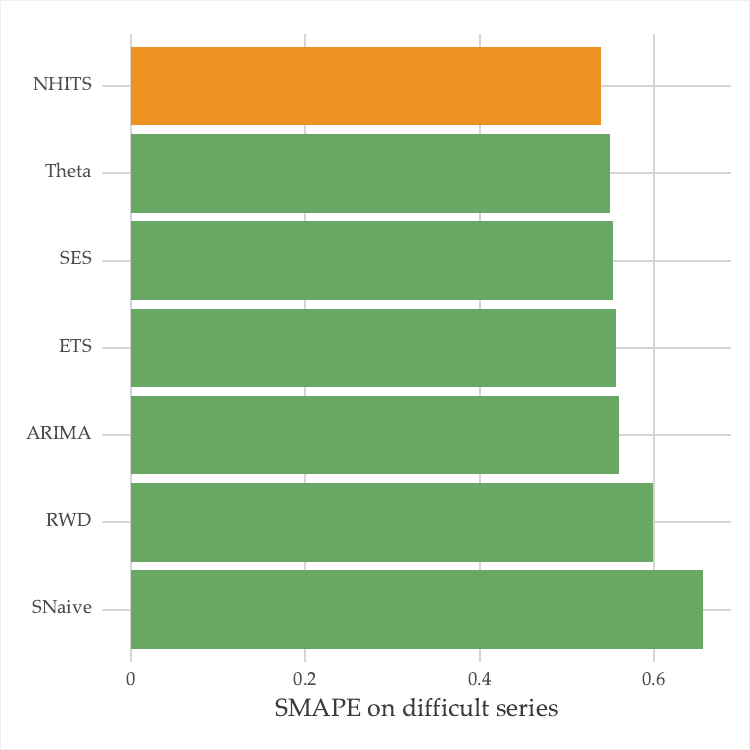} }}%
    \qquad
    \subfloat[\centering Expected shortfall]{{\includegraphics[width=.45\textwidth]{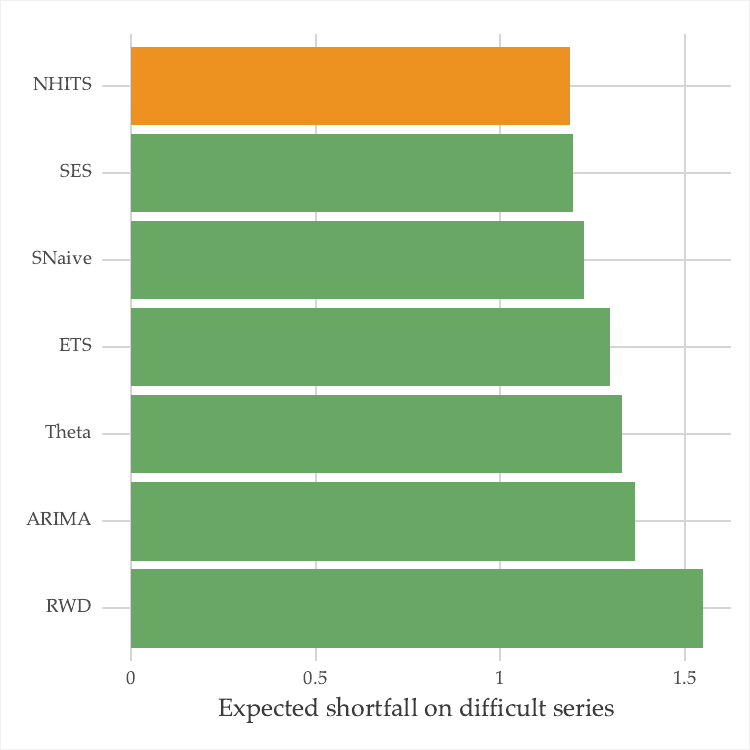} }}%
    \caption{Average SMAPE (a) and expected shortfall (b) for each model across difficult time series}%
    \label{fig:plot7}%
\end{figure}

We present the results of the repeated analysis in  Figure \ref{fig:plot7}. \texttt{NHITS} also shows the best performance in difficult problems. However, the advantage is considerably smaller relative to the results using all time series.

\subsection{Performance on anomalies}

\begin{figure}%
    \centering
    \subfloat[\centering Overall SMAPE]{{\includegraphics[width=.45\textwidth]{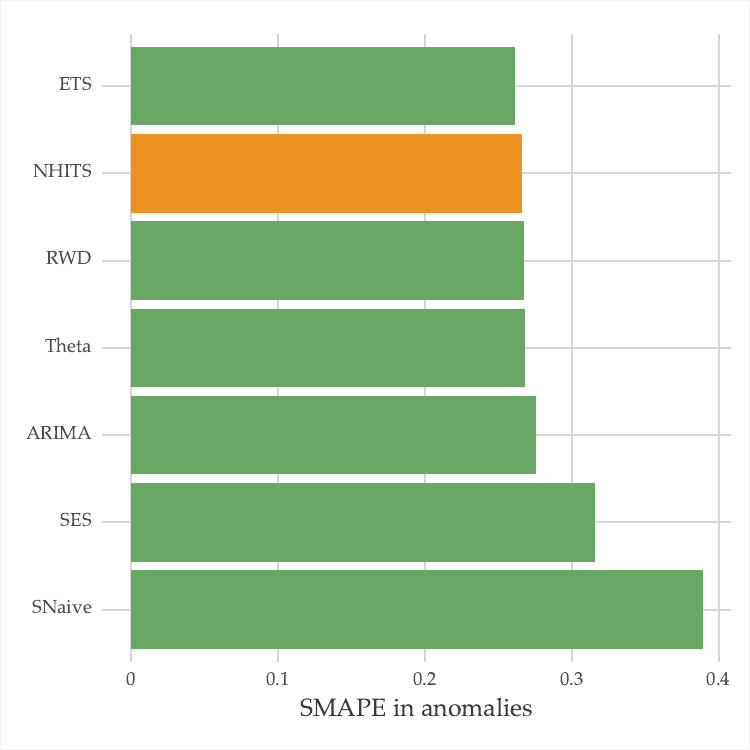} }}%
    \quad
    \subfloat[\centering Expected shortfall]{{\includegraphics[width=.45\textwidth]{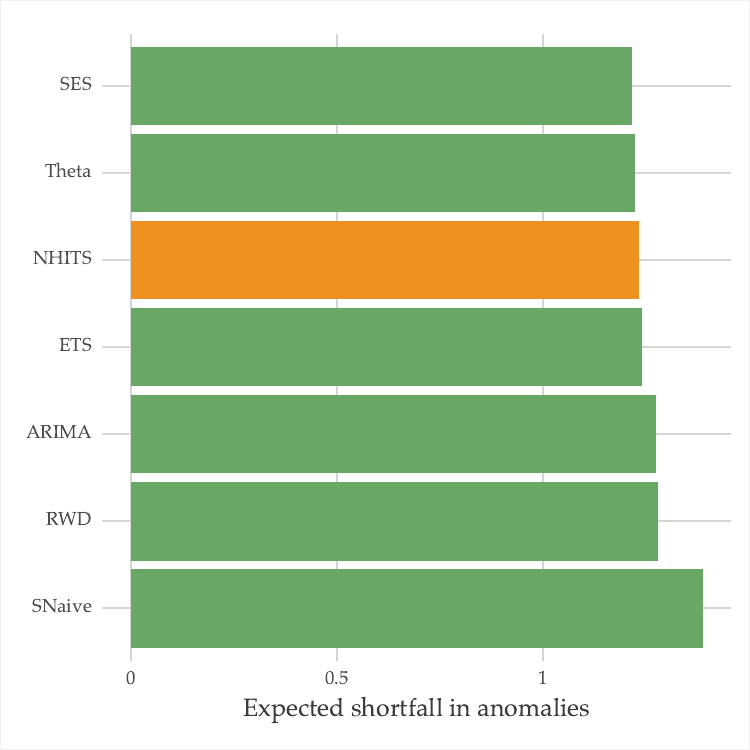}}}%
    \caption{Performance of each model in anomalous observations across all time series}%
    \label{fig:plot11}%
\end{figure}

Time series often exhibit unexpected or anomalous observations. Sometimes, these instances significantly impact the corresponding application domain, making it important to accurately forecast this type of case.

Figures \ref{fig:plot11}a and \ref{fig:plot11}b shows the performance of each model in anomalous observations across all time series. In these instances, \texttt{NHITS} is outperformed by \texttt{ETS} in terms of overall SMAPE and by \texttt{SES} and \texttt{Theta} in terms of expected shortfall.

\section{Discussion}\label{sec:discussion}

As reported in previous studies, we found that \texttt{NHITS} shows an overall better univariate forecasting performance relative to classical approaches, according to SMAPE \cite{challu2023nhits}.
However, we also discovered several factors that give a more nuanced perspective about their relative performance:
\begin{enumerate}
    \item Effect of sampling frequency: \texttt{NHITS} shows the best performance in all three sampling frequencies tested. However, \texttt{NHITS} is less competitive for time series with low sampling frequencies, such as yearly. This suggests that the effectiveness of \texttt{NHITS} may depend on the frequency at which data is collected. We note that our analysis was based on time series datasets with a monthly, quarterly, and yearly sampling frequency. This type of dataset tends to comprise many time series, but each of which is small. Notwithstanding, there is also evidence that \texttt{NHITS} shows state-of-the-art forecasting accuracy in time series with high sampling frequency \cite{challu2023nhits}.

    \item Relative performance: While \texttt{NHITS} shows better SMAPE scores overall, there is a reasonable chance that classical approaches may outperform it, even with an equivalence margin of 5\%. This implies that the superiority of \texttt{NHITS} is not guaranteed in all cases.

    \item Worst-case scenarios: In worst-case scenarios, as measured by SMAPE-based expected shortfall, \texttt{NHITS} demonstrates better performance than classical methods. This suggests that \texttt{NHITS} may be more robust or reliable relative to classical approaches.

    \item Forecasting horizon: \texttt{NHITS} is particularly suited in forecasting multiple steps ahead. This indicates that its strengths lie in long-term prediction rather than short-term forecasting. Indeed, \texttt{NHITS} was specially designed to handle long-horizon forecasting \cite{challu2023nhits}. However, previous work also reported this effect when comparing \texttt{MLPs} with \texttt{ARIMA} \cite{tang1991time}. 

    \item Difficulty of problems: The advantage of \texttt{NHITS} diminishes on difficult forecasting problems, as measured by the \texttt{SNaive} worst-case performance. This implies that the advantage of \texttt{NHITS} may vary depending on the complexity or nature of the data being analyzed.

    \item Anomalous observations: \texttt{NHITS} is outperformed by classical methods when dealing with anomalous observations. This suggests that \texttt{NHITS} may struggle with handling outliers or unexpected data points compared to classical forecasting techniques.
\end{enumerate}

Overall, these findings highlight the nuanced nature of the performance of \texttt{NHITS} compared to classical forecasting methods, with its strengths and weaknesses becoming apparent under different conditions. 
In future work, we plan to include additional perspectives to improve the characterization of the relative performance of forecasting models.

\section{Conclusions}\label{sec:conclusions}

This paper presents an extensive empirical comparison of a state-of-art deep learning forecasting method and several classical approaches for univariate time series forecasting problems.
Contrary to previous attempts at this task, we evaluate forecasting performance from different perspectives. This approach enabled a more granular analysis of the relative performance of different methods.

\texttt{NHITS} shows the overall best performance according to SMAPE, a commonly used forecasting evaluation metric. However, we found that \texttt{NHITS} is outperformed by classical approaches in a reasonable percentage of time series.
We discovered other interesting aspects, such as the varying relative performance in forecasting horizon conditions. While \texttt{NHITS} is more robust than classical approaches in terms of worst-case performance, it presents a poor performance when predicting unexpected values.
We believe that the nuanced analysis presented in this work will foster further research to develop better forecasting approaches.

\bibliographystyle{splncs04}

\end{document}